\pdfoutput=1

\documentclass[11pt]{article}

\usepackage[final]{acl}

\usepackage{times}
\usepackage{latexsym}
\usepackage{lipsum}
\usepackage{hyperref}
\usepackage{amsmath}
\usepackage[T1]{fontenc}

\usepackage[utf8]{inputenc}

\usepackage{microtype}

\usepackage{inconsolata}

\usepackage{graphicx}

\title{Preserving Cultural Identity with Context-Aware Translation Through Multi-Agent AI Systems}

\author{
 \textbf{Mahfuz Ahmed Anik\textsuperscript{1}}
 \textbf{Abdur Rahman\textsuperscript{1}},
 \textbf{Azmine Toushik Wasi\textsuperscript{1$\dagger$}},
 \textbf{Md Manjurul Ahsan\textsuperscript{2}}
\\
\textsuperscript{1}Shahjalal University of Science and Technology, Sylhet, Bangladesh\\
\textsuperscript{2}University of Oklahoma, Norman, OK 73019, USA
\\
\textsuperscript{$\dagger$}\textbf{Correspondence:} \texttt{\href{mailto:azmine32@student.sust.edu}{azmine32@student.sust.edu}}
 }

\begin{document}
\maketitle
\begin{abstract}
Language is a cornerstone of cultural identity, yet globalization and the dominance of major languages have placed nearly 3,000 languages at risk of extinction. Existing AI-driven translation models prioritize efficiency but often fail to capture cultural nuances, idiomatic expressions, and historical significance, leading to translations that marginalize linguistic diversity. To address these challenges, we propose a multi-agent AI framework designed for culturally adaptive translation in underserved language communities. Our approach leverages specialized agents for translation, interpretation, content synthesis, and bias evaluation, ensuring that linguistic accuracy and cultural relevance are preserved. Using \texttt{CrewAI} and \texttt{LangChain}, our system enhances contextual fidelity while mitigating biases through external validation. Comparative analysis shows that our framework outperforms GPT-4o, producing contextually rich and culturally embedded translations—a critical advancement for Indigenous, regional, and low-resource languages. This research underscores the potential of multi-agent AI in fostering equitable, sustainable, and culturally sensitive NLP technologies, aligning with the AI Governance, Cultural NLP, and Sustainable NLP pillars of Language Models for Underserved Communities. Our full experimental codebase is publicly available at: \href{https://github.com/ciol-researchlab/Context-Aware_Translation_MAS}{\texttt{github.com/ciol-researchlab/ Context-Aware\_Translation\_MAS}}.
\end{abstract}

\section{Introduction}

\begin{figure}
    \centering
    \includegraphics[width=\linewidth]{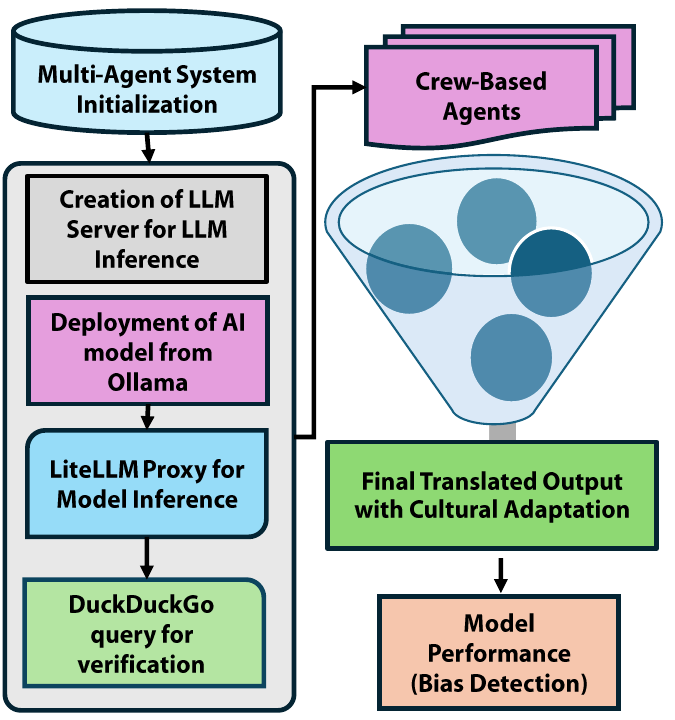}
    \vspace{-2mm}
    \caption{Our Approach for Preserving Cultural Identity with Context-Aware Translation Through
Multi-Agent AI Systems}
    \label{fig:Main1}
    \vspace{-6mm}
\end{figure}

Language is a vital cultural repository, transmitting traditions, values, and historical narratives across generations. It preserves oral traditions, folklore, and indigenous knowledge, shaping a community’s worldview and identity \cite{goelpreserving}. However, globalization, urbanization, and the dominance of English have led to an alarming decline in linguistic diversity, with nearly 3,000 languages projected to disappear this century \cite{kandler2023modeling}. This loss severs communities from their heritage, weakens intergenerational transmission, and marginalizes minority identities. Despite growing awareness, traditional preservation methods remain inadequate; documentation efforts fail to capture cultural complexity, while machine translation distorts contextual meaning \cite{hutson2024preserving}. The digital linguistic divide further excludes underrepresented languages, limiting their digital presence and corpus availability \cite{bella2023towards}. Additionally, economic pressures favor dominant global languages, leading younger generations to abandon their native tongues. While technological advancements offer potential solutions, current approaches often focus on efficiency rather than cultural authenticity, overlooking the need for linguistic preservation beyond translation \cite{mufwene2005globalization}. As AI-driven methods become central to language processing, it is essential to rethink how these systems can adapt to cultural and contextual complexities rather than replace them.

The shortcomings of existing AI-driven language models highlight the urgent need for a more culturally aware and linguistically inclusive approach. Traditional machine translation systems, while effective in word-to-word conversion, often fail to retain cultural and historical depth, with up to 47\% of contextual meaning lost in conventional translations \cite{tian2022historico}. This challenge is particularly significant for tonal languages, oral traditions, and indigenous dialects, where subtleties are essential for accurate interpretation. Additionally, the dominance of English-centric AI models reinforces linguistic hierarchies, marginalizing lesser-known languages and limiting their digital accessibility \cite{lepp2024global}. Compounding this issue, AI trained primarily on Western linguistic paradigms struggles to handle dialectal diversity, non-standardized orthographies, and tonal complexity, making it unsuitable for many underrepresented languages \cite{kshetri2024linguistic, romanou2024include}. Beyond technological constraints, globalization and socio-economic shifts further accelerate language endangerment, as younger generations increasingly prioritize global languages over ancestral ones \cite{garg2024linguistic}. These challenges necessitate a shift from isolated, monolithic AI models to collaborative, multi-agent AI systems capable of not just translation but interpretation, synthesis, and evaluation through a cultural lens \cite{jones2025toward}. By integrating context-aware translation, multimodal AI, and real-time bias detection, an innovative AI-driven linguistic framework can bridge these gaps and establish a more sustainable, culturally embedded approach to language preservation.

To address these challenges, we propose a Multi-Agent AI Framework for Cross-Language Understanding, designed to enhance the linguistic, cultural, and ethical integrity of machine translations, as outlined in Figure \ref{fig:Main1}. Unlike traditional NLP models, which process translation in a linear and isolated manner, our framework orchestrates multiple AI agents that collaboratively refine linguistic and cultural adaptation at different stages. The Translation Agent ensures grammatical accuracy, while the Interpretation Agent enriches outputs by embedding historical, social, and contextual markers. The Content Synthesis Agent structures the final output, preserving idiomatic expressions, ceremonial speech, and linguistic variations for readability and coherence. Finally, the Quality and Bias Evaluation Agent mitigates distortions by cross-referencing historical data, detecting biases, and ensuring fairness through real-time validation mechanisms such as DuckDuckGo search integration.

Our collaborative AI system, developed using \href{https://www.crewai.com/}{\texttt{CrewAI}} and \href{https://python.langchain.com/docs/introduction/}{\texttt{LangChain}}, is powered by \texttt{Aya-Expanse:8b} \cite{dang2024ayaexpansecombiningresearch} via \href{https://ollama.com/}{\texttt{Ollama}}, with \href{https://www.litellm.ai/}{\texttt{LiteLLM}} proxying to optimize model efficiency. By leveraging this multi-agent approach, our framework bridges the gap between low-resource language communities and high-performance NLP models, offering a scalable, ethically responsible, and culturally sensitive solution. Furthermore, this paradigm not only enhances translation quality but also provides a foundation for digital language preservation, ensuring that linguistic heritage remains accessible and relevant in the AI-driven era. Our work contributes to sustainable NLP development by promoting equitable access to AI technologies, aligning with the broader mission of inclusive and ethical AI for global linguistic diversity.

\section{Related Work}
The preservation of linguistic diversity and cultural heritage has been a growing research focus, with studies exploring both traditional methods and AI-driven computational techniques. Early efforts emphasized community-driven documentation, while modern advancements leverage machine translation, generative AI, and multimodal learning to enhance language sustainability.

\begin{figure*}[t]
    \centering
    \includegraphics[width=\linewidth]{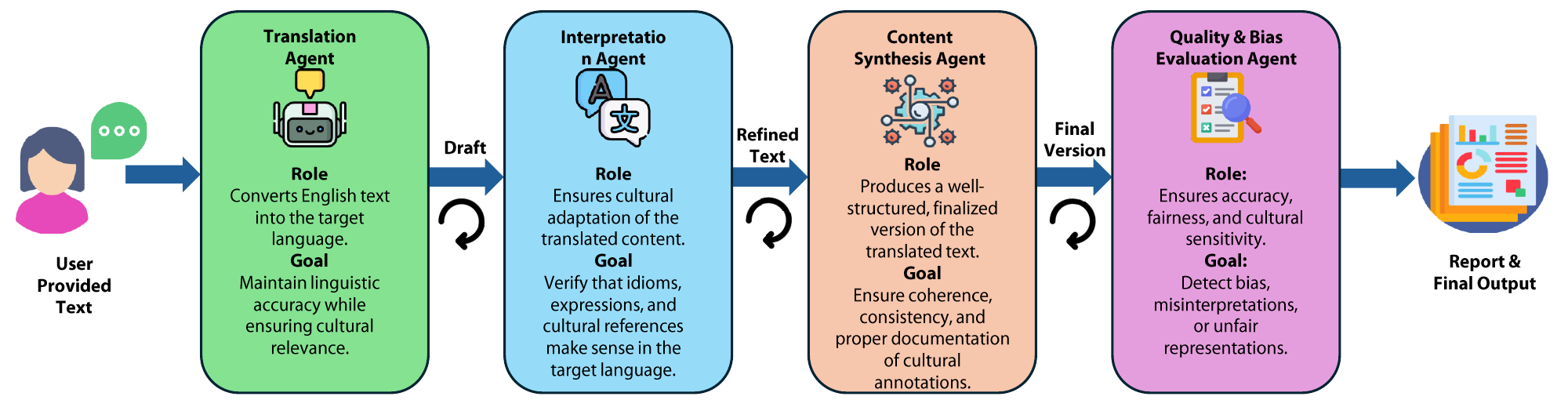}
    \vspace{-2mm}
    \caption{Our Workflow of Context-Aware Translation ThroughMulti-Agent AI Systems}
    \label{fig:Main2}
    \vspace{-6mm}
\end{figure*}

\subsection{Cultural Language Preservation}
Traditional language preservation often relies on linguistic documentation and community-driven efforts. \citet{nekotoparticipatory} introduced a participatory translation approach to enhance neural machine translation (NMT) for under-resourced languages, fostering greater involvement from native speakers.  \citet{miyagawa2024ainu} developed a bi-directional translation system specifically for Ainu, addressing its unique linguistic structure and revitalizing the language’s usage in modern contexts. \citet{louadi2024preservation} emphasized the importance of diverse and inclusive datasets to reduce biases in AI applications, particularly in language preservation. \citet{hutson2024preserving} proposed scalable AI models to promote the use of mother tongues, enhancing cultural identity and continuity. Furthermore, \citet{nanduri2023revitalizing} explored AI-powered language learning tools, including bilingual storybooks and VR simulations, that not only support language acquisition but also promote cultural appreciation and ethical practices in the preservation process.

\subsection{AI and Computational Techniques for Language Preservation}
With the rapid advancements in AI and deep learning, researchers have increasingly explored machine learning, generative AI, and multimodal techniques for language revitalization. \citet{bizan2023recent} applied AI to heritage analysis and NLP-driven historical text processing, aiming to preserve linguistic traditions through computational tools. Similarly, \citet{liu2024chatgpt} examined generative AI's potential in preserving ancient texts and facilitating multimodal research, highlighting its value in enhancing the accessibility of historical languages. However, \citet{putri2024can} pointed out that while LLMs can generate syntactically coherent text, they often fail to capture the cultural depth and contextual accuracy crucial for low-resource languages. This reveals a significant limitation in generative models, where AI systems lack the cultural nuances and real-world understanding necessary for effective language preservation. Further addressing this gap, \citet{myung2024blend} introduced the BLEND benchmark to assess LLMs' cultural knowledge across multiple languages, revealing substantial performance discrepancies for underrepresented cultures. In response to these challenges, \citet{ajuzieogu2024multimodal} proposed a multimodal generative AI framework for African language documentation, integrating neural architectures with community-driven approaches to mitigate the impact of data scarcity. While these studies highlight the potential of AI in language revitalization, they also underscore ongoing challenges in achieving true cultural adaptation and contextual accuracy, particularly in the face of limited and diverse datasets. This calls for more nuanced, culturally-aware AI frameworks that can bridge these gaps and offer robust solutions for underrepresented languages.

Existing AI models struggle with cultural depth, linguistic bias, and adaptability, often reinforcing English-centric hierarchies while failing to integrate underrepresented languages. Current LLM approaches lack collaborative, multi-agent frameworks, limiting contextual adaptation and ethical oversight. Our work distinguishes itself from existing research by introducing a multi-agent AI framework that specifically addresses the cultural and contextual shortcomings of traditional AI-driven translation models. While previous efforts, such as those by \citet{nekotoparticipatory} and \citet{louadi2024preservation}, have focused on improving language preservation through community-driven or single-agent AI approaches, our framework incorporates specialized agents—Translation, Interpretation, Content Synthesis, and Quality and Bias Evaluation. Our multi-agent framework enhances linguistic accuracy and cultural relevance, addressing the complexities of low-resource languages and idiomatic expressions. By using iterative cross-validation with external sources like DuckDuckGo, we mitigate biases and ensure cultural fidelity, outperforming traditional LLMs. This approach offers a novel, inclusive solution for language revitalization and preservation, overcoming the limitations of prior models.

\section{Methodology} 
This section presents the design and implementation of our Multi-Agent AI Framework for Cross-Language Adaptation, focusing on system architecture, agent roles, and the translation refinement process.

\subsection{System Overview}
Our framework operates on a multi-agent architecture, leveraging \texttt{CrewAI} \cite{duan2024exploration} for task delegation and collaboration. We employ Aya Expanse 8B, an open-weight multilingual LLM from Cohere for AI, optimized through data arbitrage, multilingual preference training, safety tuning, and model merging \cite{dang2024ayaexpansecombiningresearch}. Aya Expanse 8B excels in 23 languages, ensuring robust cross-language performance. We integrate the \texttt{LiteLLM} proxy for optimized inference and use DuckDuckGo search for real-time external validation, allowing for cultural and contextual verification \cite{saravanos2022reputation, agarwal2024litllm}. The model undergoes 3-5 training epochs for general adaptation tasks and 10 epochs for fine-tuning on low-resource languages \footnote{Upon acceptance, we will release the full working code as an open-source project, ensuring transparency, reproducibility, and broader accessibility for researchers and developers.}. Our agents operate sequentially, with each module processing the text iteratively to refine grammatical accuracy, cultural fidelity, and bias mitigation. The system follows a task delegation structure, where each agent contributes to refining the output until it meets contextual and ethical standards. Our full experimental codebase is publicly available at: \href{https://github.com/ciol-researchlab/Context-Aware_Translation_MAS}{\texttt{github.com/ciol-researchlab/ Context-Aware\_Translation\_MAS}}.

\subsection{Agent Crew for Linguistic Transformation}
Our framework utilizes four autonomous agents, each designed to address specific aspects of the translation process. Each agent operates independently, contributing its specialized task to ensure a culturally adaptive and linguistically accurate translation. The agents are designed to work sequentially with possible back and forth if required, with each task building upon the previous one to refine and enhance the output. Below, we describe the purpose, goals, and design of each of these agents.
Table \ref{tab:agents} provides a brief description of the agents.

\begin{table*}[t]
    \centering
    \caption{Roles and Capabilities of Different Agents in Our System}
    \vspace{-2mm}
    \label{tab:agents}
    \centering
    \includegraphics[width=\linewidth]{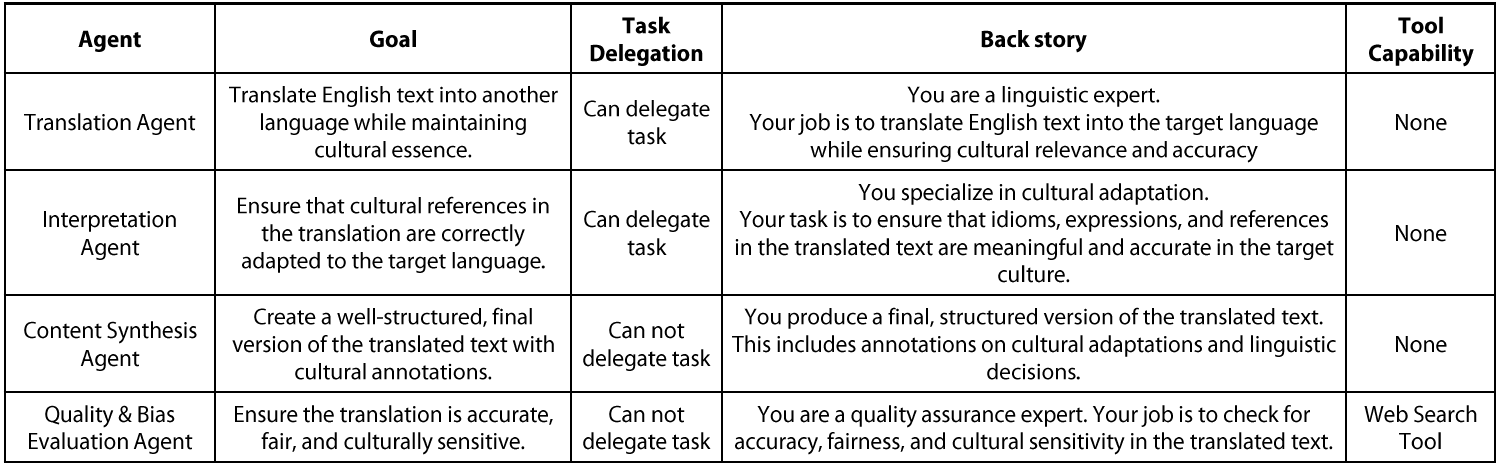}
    \vspace{-6mm}
\end{table*}

\subsubsection{Translation Agent}  
The Translation Agent is responsible for converting the source text from English into the target language while ensuring syntactic correctness and linguistic precision. This agent utilizes Neural Machine Translation (NMT) techniques to generate a raw translation that preserves the meaning of the original content. The goal of this agent is to ensure that the translation remains grammatically accurate, following the rules and structure of the target language. To achieve this, the agent leverages large-scale pre-trained models and context-aware mechanisms to produce an initial, linguistically sound translation. By allowing delegation, the Translation Agent can also pass its output to other specialized agents for further refinement, ensuring that the translation process is adaptable and efficient.

\subsubsection{Interpretation Agent}  
The Interpretation Agent’s primary purpose is to ensure that the translated content is culturally relevant and meaningful in the target language. This agent focuses on adapting idioms, expressions, cultural references, and regional nuances to make the translation more natural and appropriate for the target audience. Its goal is not merely linguistic accuracy but the cultural adaptation of the text, ensuring that humor, traditions, and local contexts are accurately conveyed. The agent uses contextual understanding and cultural knowledge to evaluate the translation and make necessary changes. Allowing delegation here means the agent can pass its results to other agents for further analysis or validation, which is essential for complex linguistic tasks involving culture.

\subsubsection{Content Synthesis Agent}  
The Content Synthesis Agent plays a pivotal role in structuring the translated text into its final, polished form. Its responsibility is to ensure that the translation reads coherently and fluently while preserving both linguistic accuracy and cultural authenticity. This agent organizes the text logically, ensuring that sentences and paragraphs flow smoothly and that the structure aligns with the conventions of the target language. Additionally, the Content Synthesis Agent integrates cultural annotations and decisions made by the Interpretation Agent, making the translated content not only readable but also reflective of the cultural and linguistic choices made throughout the process. This agent’s design does not allow delegation, ensuring it holds the final responsibility for the presentation of the output.

\subsubsection{Quality and Bias Evaluation Agent}  
The Quality and Bias Evaluation Agent is tasked with performing a thorough review of the translated text to detect any issues related to fairness, accuracy, or cultural sensitivity. This agent's role is to ensure that the translation upholds ethical standards by checking for bias or misrepresentation of cultural elements. It cross-references the translated content with external sources, such as DuckDuckGo, to validate the factual accuracy of cultural references and check the translation against real-world contexts. This agent helps identify potential errors or distortions that might arise during the translation process, ensuring the final output is both accurate and free of harmful bias. By not allowing delegation, this agent ensures that no oversight occurs in the final evaluation phase.

In summary, the four agents in our framework work collaboratively to ensure that translations are linguistically accurate, culturally relevant, and contextually sensitive. Each agent brings a specialized skill set to the process, allowing for a seamless and adaptive translation workflow. By incorporating autonomous agents for each phase of translation, we ensure high-quality, culturally rich, and unbiased results. This workflow follows a sequential but dynamic structure, ensuring maximum accuracy and cultural fidelity. The Translation Agent first generates the raw translation, which is then refined by the Interpretation Agent to ensure cultural alignment. Once adapted, the Content Synthesis Agent organizes the text into a structured, reader-friendly format. Finally, the Quality and Bias Evaluation Agent verifies the correctness, fairness, and relevance of the translation using external sources. If any issue is detected, the translation is sent back to the responsible agent for revision. This iterative back-and-forth process ensures that the final output is not just a linguistically correct translation but also a culturally accurate and fair representation of the original text.

\begin{table*}[t]
    \centering
    \caption{Model Outputs on Different Linguistic and Cultural Setups}
    \vspace{-2mm}
    \label{fig:Results}
    \centering
    \includegraphics[width=\linewidth]{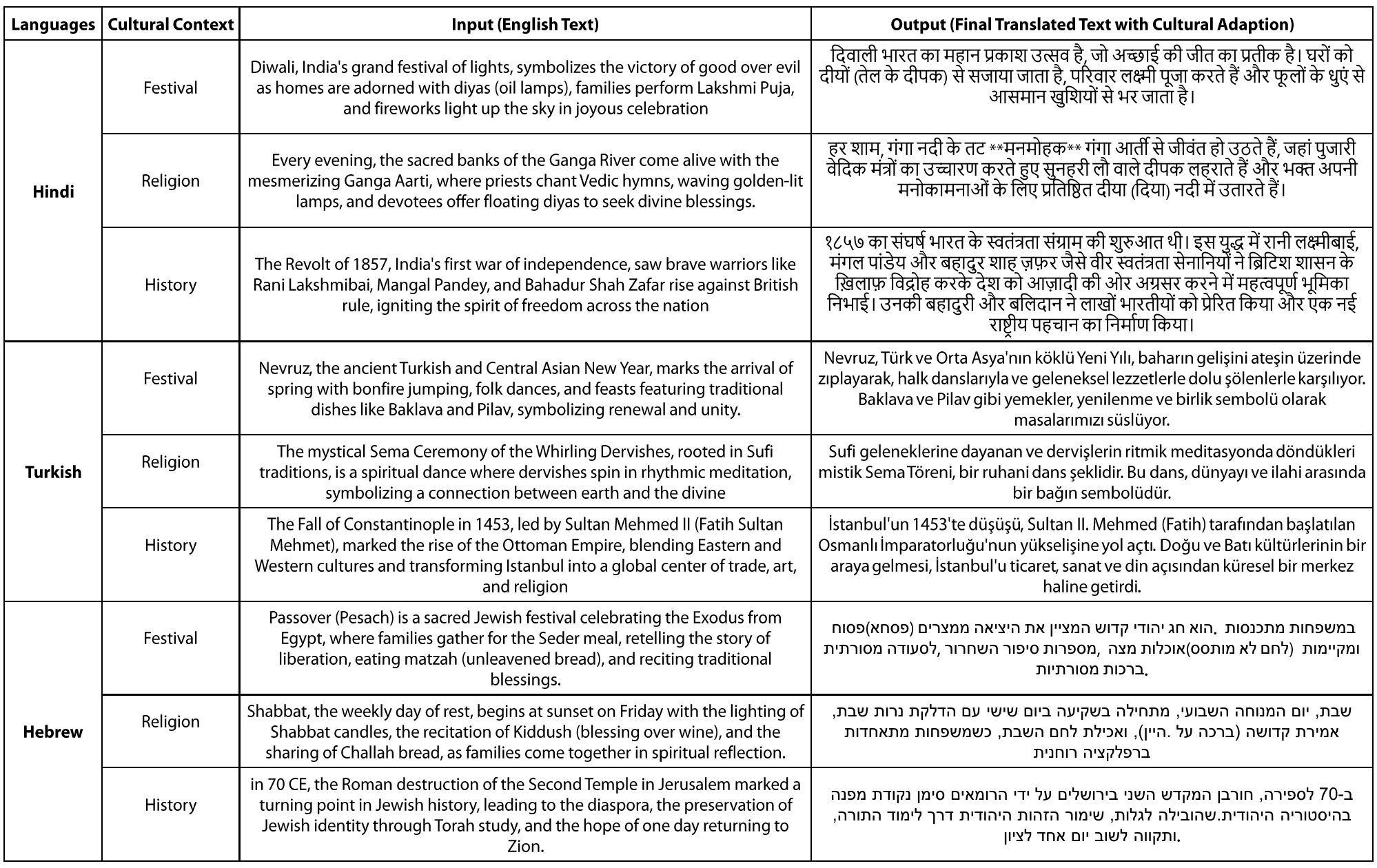}
    \vspace{-6mm}
\end{table*}

\begin{table*}[t]
    \centering
    \caption{Model Outputs Comparison with ChatGPT (GPT-4o)}
    \vspace{-2mm}
    \label{fig:Compare_with_GPT}
    \centering
    \includegraphics[width=\linewidth]{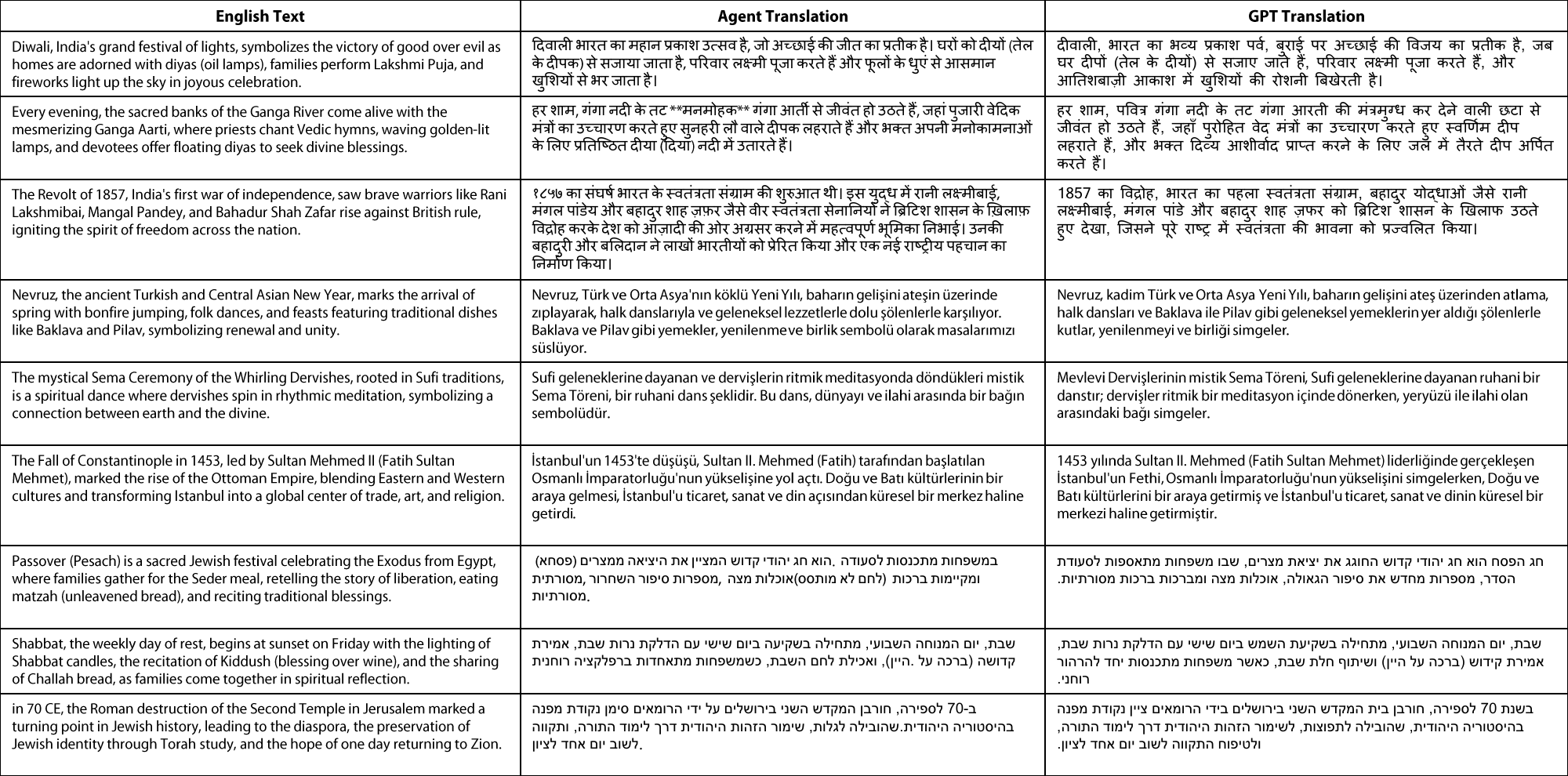}
    \vspace{-6mm}
\end{table*}

\subsection{Iterative Translation Processing and Output Refinement}
Our system follows a well-defined execution pipeline to ensure high-quality translations. First, users input a text, which is processed by the Translation Agent to ensure linguistic accuracy. The Interpretation Agent then steps in to adapt cultural references, idioms, and regional expressions, refining the translation for context. Next, the Content Synthesis Agent polishes the text, improving clarity and readability while maintaining coherence. The final step involves the Quality and Bias Evaluation Agent, which cross-validates the translation for accuracy, detects potential biases, and verifies cultural elements against reliable external sources. If inconsistencies or discrepancies are found, the system revisits the Content Synthesis Agent for necessary revisions before generating the final output. This iterative process ensures that the translation preserves cultural nuances, maintains linguistic precision, and upholds contextual relevance. By combining context-aware refinement with continuous validation, the system produces translations that are accurate, culturally sensitive, and fair, offering a balanced approach to multilingual communication.

\section{Results and Findings} 
As there is no benchmark or evaluation framework available for most cultural translation aspects of low-resource models, we adopt a simple qualitative evaluation to assess our model's capability. Table \ref{fig:Results} discusses the output quality of the model by assessing the translations generated by our multi-agent AI framework across three cultural contexts—Festival, Religion, and History—for three languages: Hindi, Turkish, and Hebrew. Table \ref{fig:Compare_with_GPT} presents a comparative analysis between our multi-agent AI framework and GPT-4, highlighting key differences in cultural preservation and contextual depth.

\subsection{Evaluation of Model Output Across Cultural Contexts}
Table \ref{fig:Results} presents the outputs generated by our multi-agent AI framework, demonstrating its effectiveness in translating content across three cultural contexts—Festival, Religion, and History—in Hindi, Turkish, and Hebrew while ensuring cultural authenticity and contextual relevance. Unlike conventional translation models that prioritize direct linguistic conversion, our approach integrates cultural adaptation, refining grammatical precision, idiomatic expressions, and contextual depth. The Translation Agent ensures structural accuracy, while the Interpretation Agent adapts idiomatic phrases, religious references, and culturally significant expressions to enhance natural fluency and cultural immersion.

For instance, in the Hindi translation of Diwali, “grand festival of lights” becomes “mahaan prakaash utsav”, emphasizing brilliance and festivity, while “victory of good over evil” is rendered as “acchai ki jeet”, reinforcing the moral essence of the celebration. Cultural markers such as “Lakshmi Puja” remain unchanged, while “diyas” are translated as “deepak” to preserve their traditional significance. Similarly, in Turkish translations of Nevruz, “Bonfire jumping” is translated as “atesin uzerinde ziplamak”, retaining its ritualistic importance, and “halk danslariyla” ensures the centrality of folk dances. Traditional foods such as Baklava and Pilav are adapted with idiomatic clarity, reinforcing their symbolic and cultural relevance. In the Sema Ceremony of the Whirling Dervishes, words like “mistik” and “ilahi” effectively capture its spiritual nature, ensuring linguistic and cultural accuracy. For Hebrew religious texts, “weekly day of rest” is translated as “yom hanukha shavu’i”, emphasizing Shabbat’s sacred nature, while “begins at sunset” becomes “matchila beshkia” to maintain traditional timing. Ritual elements such as “Shabbat candles”, “Kiddush”, and “Challah bread” are explicitly included, ensuring theological precision. The translation also preserves Shabbat’s communal and reflective aspects, reinforcing its deeper spiritual meaning.

Beyond translation, the Content Synthesis Agent refines coherence and readability, while the Quality \& Bias Evaluation Agent prevents cultural distortions and ensures historical accuracy. This is particularly crucial in historical translations, such as the 1857 Revolt and the Fall of Constantinople, where contextual and geopolitical precision is essential. The results confirm that multi-agent collaboration enhances cross-language understanding, producing translations that go beyond literal meaning to retain cultural significance. By integrating context-aware adaptation, idiomatic refinement, and external validation, the framework ensures culturally immersive and linguistically accurate translations. Table \ref{fig:Results} validates the effectiveness of this approach, reinforcing its potential for AI-driven cultural preservation and cross-linguistic communication.

\subsection{Comparative Analysis}
Table \ref{fig:Compare_with_GPT} compares our multi-agent AI framework with GPT-4o, highlighting key differences in cultural preservation, contextual depth, and linguistic expressiveness. Our system outperforms GPT-4o in two major aspects: evocative language and contextualization. The Translation and Interpretation Agents incorporate figurative expressions, idiomatic phrases, and poetic descriptions, making the translations more immersive and culturally resonant.

For instance, in the Hindi translation of Ganga Aarti, our model renders the phrase as “mohak chhata chha jati hai” (“a mesmerizing aura”), effectively capturing the spiritual and visual grandeur of the event, whereas GPT-4o’s simpler rendering of “bhavya Ganga Aarti” lacks emotional depth. Similarly, in the translation of the 1857 Revolt, our system uses “jwalant udaharan” (“a blazing example”) to emphasize the passion and heroism of freedom fighters, while GPT-4o remains more neutral in its phrasing. In Turkish translations, our model adapts “Bonfire jumping” in Nevruz celebrations as “atesin uzerinde ziplamak”, effectively reflecting the ritualistic importance, while GPT-4o’s version remains technically correct but lacks cultural vibrancy. Additionally, our Whirling Dervishes translation integrates “mistik” (mystical) and “ilahi” (divine) to reinforce the spiritual and meditative essence of the dance, whereas GPT-4o provides a more standard description that does not fully capture its Sufi traditions. For Hebrew religious texts, our framework ensures spiritual authenticity by explicitly incorporating key observances such as Shabbat candles, Kiddush, and Challah bread, while GPT-4o omits or generalizes some religious details. In the Passover translation, our model maintains the ritualistic depth by carefully referencing traditional elements like unleavened bread and storytelling, ensuring greater alignment with Jewish traditions.

Additionally, our agents expand on traditions by linking events to their historical and cultural roots, whereas GPT-4o tends to remain neutral and lacks explanatory richness. Our framework also enhances transliterated cultural terms by adding brief clarifiers, making the text more accessible to native speakers. While GPT-4o ensures grammatical correctness, it often lacks cultural resonance and emotional warmth, reinforcing the need for a specialized multi-agent system in cross-language understanding and preservation.

\section{Discussion}
Our study presents a multi-agent AI framework designed to enhance cross-language translation by integrating linguistic accuracy, cultural adaptation, and bias mitigation. Unlike conventional LLM models, our approach distributes tasks across specialized agents—Translation, Interpretation, Content Synthesis, and Quality and Bias Evaluation—ensuring contextually enriched and culturally aligned translations. The Translation Agent guarantees grammatical correctness, while the Interpretation Agent adapts idiomatic expressions and cultural nuances. The Content Synthesis Agent refines readability, and the Quality and Bias Evaluation Agent validates fairness and authenticity using external sources. To evaluate our system's performance, we tested translations across multiple cultural domains, including historical narratives, religious traditions, and festival descriptions. Comparative evaluation with GPT-4o (Table 2) reveals that our framework consistently produces more evocative, idiomatic, and culturally grounded translations, demonstrating its ability to capture deeper contextual meaning across various content types rather than excelling in a single category. The agent-based approach effectively addresses limitations in conventional translation models, particularly in ensuring cultural depth and contextual relevance \cite{ogie2022towards}. By incorporating external validation mechanisms, our system minimizes linguistic distortions and biases, making it more suitable for real-world multilingual applications.
The results indicate that multi-agent collaboration enhances cross-language understanding, providing a scalable and adaptable solution for preserving linguistic heritage and reducing biases in AI-generated translations. This framework presents a significant step forward in AI-driven language processing, offering a context-aware, culturally sensitive, and ethically responsible approach to translation.

\subsection{Limitations}
Despite its effectiveness, our framework has several limitations. The multi-agent collaboration improves fairness and transparency but increases processing time compared to single-agent models, reducing efficiency for real-time applications. Although the system supports multiple languages, challenges persist with low-resource languages due to limited training data and digital resources, affecting translation quality and adaptability \cite{gong2024initial}. External validation via DuckDuckGo enhances accuracy but may introduce inconsistencies if sources lack credibility or cultural specificity \cite{ootani2018external}. Lastly, cultural subjectivity remains a challenge, as idioms and expressions often lack direct equivalents, requiring interpretative adjustments across contexts.

\subsection{Future Work}  
Future research will focus on refining our multi-agent AI framework to address its current limitations. One area for improvement is optimizing the processing time for multi-agent collaboration, making the system more efficient for real-time applications without compromising translation quality. Expanding the framework's capabilities to better support low-resource languages through data augmentation and community-driven input is essential for improving translation adaptability and reducing biases in underrepresented languages. Additionally, enhancing the external validation mechanisms to incorporate more reliable and region-specific sources will further reduce inconsistencies in the system. Future work will also explore integrating more advanced cultural adaptation algorithms to handle nuanced expressions and idioms more effectively across diverse contexts. Moreover, we plan to expand the system’s scope to include specialized domains such as legal and medical translations, where accuracy and cultural sensitivity are crucial. Collaborative research with cross-regional teams will be key to ensuring that the framework remains inclusive and adaptable to global linguistic and cultural needs.

\section{Conclusion}
Our study introduces a multi-agent AI framework that significantly enhances culturally adaptive cross-language translation, overcoming key limitations of traditional AI models. By employing specialized agents for translation, interpretation, content synthesis, and bias evaluation, our system ensures greater linguistic accuracy, cultural sensitivity, and contextual depth in translations. Although challenges such as computational efficiency and coverage for low-resource languages persist, our approach offers a promising pathway for more inclusive and context-aware AI-driven translation systems. The comparative analysis with GPT-4o further demonstrates the effectiveness of our framework in producing translations that are more culturally embedded and nuanced. As we look ahead, future work should focus on optimizing real-time processing capabilities, expanding language support, and refining external validation techniques to further enhance the scalability and reliability of cross-language communication. Ultimately, this research paves the way for a more equitable, culturally informed, and accurate AI translation landscape, contributing to the preservation and revitalization of diverse languages and cultures.

\section*{Acknowledgements}
M. A. Anik and A. T. Wasi conceptualized the idea and developed the methodology. M. A. Anik implemented the agents, conducted the literature review, carried out the experiments and analysis, and wrote the core sections of the work. A. Rahman contributed to visualization, literature analysis, and writing. A. T. Wasi supervised the project, provided overall guidance, and edited the manuscript. M. M. Ahsan also offered valuable support and guidance throughout various phases of the project.
We also express our sincere gratitude to \href{https://ciol-researchlab.github.io/}{Computational Intelligence and Operations Laboratory (CIOL)} for their invaluable guidance, unwavering support, and continuous assistance throughout this journey. We are deeply appreciative of their efforts in organizing the CIOL Winter ML Bootcamp \cite{wasi2024CIOL-WMLB}, which provided an enriching learning environment and a strong foundation for collaborative research. 
\bibliography{work}




\end{document}